\documentclass[runningheads]{llncs}
\usepackage[T1]{fontenc}
\usepackage{graphicx}
\usepackage{amsmath}
\usepackage{xcolor}
\usepackage{hyperref}
\begin{document}
\title{Biochemical Prostate Cancer Recurrence Prediction: Thinking Fast \& Slow\\
%{\footnotesize \textsuperscript{*}Note: Sub-titles are not captured for https://ieeexplore.ieee.org  and
%should not be used}
%\thanks{Identify applicable funding agency here. If none, delete this.}
}
%\titlerunning{Abbreviated paper title}
% If the paper title is too long for the running head, you can set
% an abbreviated paper title here
%
\author{Suhang You*\inst{1} \and
Sanyukta Adap\inst{1} \and
Siddhesh Thakur\inst{1} \and
Bhakti Baheti\inst{1} \and
Spyridon Bakas*\inst{1} }
\authorrunning{S.You, et al.}
% First names are abbreviated in the running head.
% If there are more than two authors, 'et al.' is used.
%
\institute{Division of Computational Pathology, Department of Pathology \& Laboratory Medicine, Indiana University School of Medicine, Indianapolis, IN, USA\\ 
* Corresponding authors: jaydyou@iu.edu, spbakas@iu.edu} 
\maketitle              % typeset the header of the contribution
\begin{abstract}
Time to biochemical recurrence in prostate cancer is essential for prognostic monitoring of the progression of patients after prostatectomy, which assesses the efficacy of the surgery. In this work, we proposed to leverage multiple instance learning through a two-stage ``thinking fast \& slow'' strategy for the time to recurrence (TTR) prediction. The first (``thinking fast'') stage finds the most relevant WSI area for biochemical recurrence and the second (``thinking slow'') stage leverages higher resolution patches to predict TTR. Our approach reveals a mean C-index ($Ci$) of 0.733 ($\theta=0.059$) on our internal validation and $Ci=0.603$ on the LEOPARD challenge validation set. Post hoc attention visualization shows that the most attentive area contributes to the TTR prediction.

\end{abstract}
\section{Introduction}
%\textcolor{blue}{Very briefly (1-2 paragraphs) describe a problem and motivation for the approach.}
% 1 paragraph: Introduce the clinical problem
In 2020, more than 10 million new male cancer cases were diagnosed, with prostate cancer (PC) ranking second to lung cancer~\cite{sung2021global}. Currently, PC clinical treatment relies on prostatectomy targeting prolonged life expectancy. However, up to 40\% of PC patients would experience biochemical recurrence of the prostate-specific antigen %(at the levels of 0.2 ng/mL and onward) 
within 10 years~\cite{roehl2004cancer,freedland2005risk,stephenson2006defining}.%, and may develop metastasis~\cite{pound1999natural,boorjian2011long}. 
The Gleason score~\cite{gleason1974prediction} has been ranking PC on different risk grades, based on morphological features, albeit its limitations lead to recurrence rate differences within the same grade~\cite{epstein2016contemporary}.

 % 1 paragraph: Introduce how the literature has been tackling this problem so far and identify the gap we will be covering
 % 1 paragraph: Introduce MIL and a few applications
%To improve biochemical recurrence prediction, 
Recently deep learning methods~\cite{pinckaers2022predicting,eminaga2024artificial} have targeted superior biochemical recurrence prediction to the Gleason score, relying on the analysis of digitized histological images of tissue microarrays, rather than whole slide images (WSIs).%, which are extremely large both in storage (e.g., several gigabytes) and spatial resolution (e.g., $100,000\times100,000$). 
A common solution to analyze WSIs is by partitioning them into smaller patches, notwithstanding the challenge of obtaining patch-level annotations. Along these lines, multiple instance learning (MIL)~\cite{carbonneau2018multiple} has become prominent in computational pathology for many applications~\cite{gadermayr2024multiple}, as it encapsulates features from individual patches of the same WSI as a bag~\cite{sivic2008efficient}, reducing the patch-level labeling requirement and transforming it into a weakly-supervised learning problem with known bag/WSI-level labels. Direct risk prediction has been proposed in~\cite{katzman2018deepsurv} by modeling a Cox layer and recently advanced with MIL in~\cite{yao2020whole}, which groups the extracted patch-level features with K-means to improve patch sampling.

% 1 paragraph: Describe what we propose in this work and the following sections.
Motivated by this recent literature, here we propose a two-stage MIL regression approach to tackle the task of predicting biochemical recurrence in prostate cancer, as part of the LEarning biOchemical Prostate cAncer Recurrence from histopathology sliDes (LEOPARD) Challenge 2024~\cite{4tmviia9ll}. The proposed two-stage approach follows a ``thinking fast \& slow'' strategy, towards improving the patch sampling/pooling and targeting inference efficiency. Specifically, the $1^{st}$ stage aims to rapidly localize the most important WSI area, and the $2^{nd}$ stage leverages these important patches and focuses on selecting the most attentive features to predict TTR.% In the following sections, we first discuss the data we used during experimentation, and detail our two-stage ``thinking fast'' and ``thinking slow'' strategy. We then present our experimental design and the results from both internal data splits and the validation leaderboard produced during our participation in the LEOPARD challenge. Finally, we further explore the interpretability of our proposed method, discuss its limitations, and recommend future directions. 

\section{Material}\label{sec:material}
%\textcolor{blue}{The participants must indicate if any external data or pre-trained weights were used. There is no need to describe the LEOPARD dataset, however, if there are any data-related details or techniques you used -  this is the section to describe them, e.g. tuning data split, using additional public data, using pre-trained weights ...  }
We developed our model using the LEOPARD challenge training set (508 cases). We used all training data for our $2^{nd}$ stage (Sec.~\ref{sec:stage_2}), and excluded 30\% for our $1^{st}$ stage (Sec.~\ref{sec:stage_1}) by setting the time threshold $T=1.65$, where cases with $e_i = 0$ (no recurrence) and $t_i \text{ (follow-up years) } < T$ are excluded. For both stages, the split ratios for training, validation, and testing were 64\%:16\%:20\%.
% We then use $e_i$ as the ground truth of recurrence for the selected WSIs. 
UNI was used as the feature extractor, including its pre-trained weights~\cite{chen2024towards}.

Our final model was submitted to the LEOPARD Challenge validation and testing phase. The validation set comprised 49 cases from `Radboud' and 50 cases from external sources. The testing set was hidden from challenge participants.

\section{Methods}
%\textcolor{blue}{describe in detail the steps of your methodology: pre-processing, data filtering/denoising, model/s, model/s training strategy, loss, model/s tuning strategy, experiments... In case any techniques for improving domain generalization were used, be sure to mention those.}

\begin{figure*}[h]
  \centering
  \includegraphics[width=0.95\textwidth]{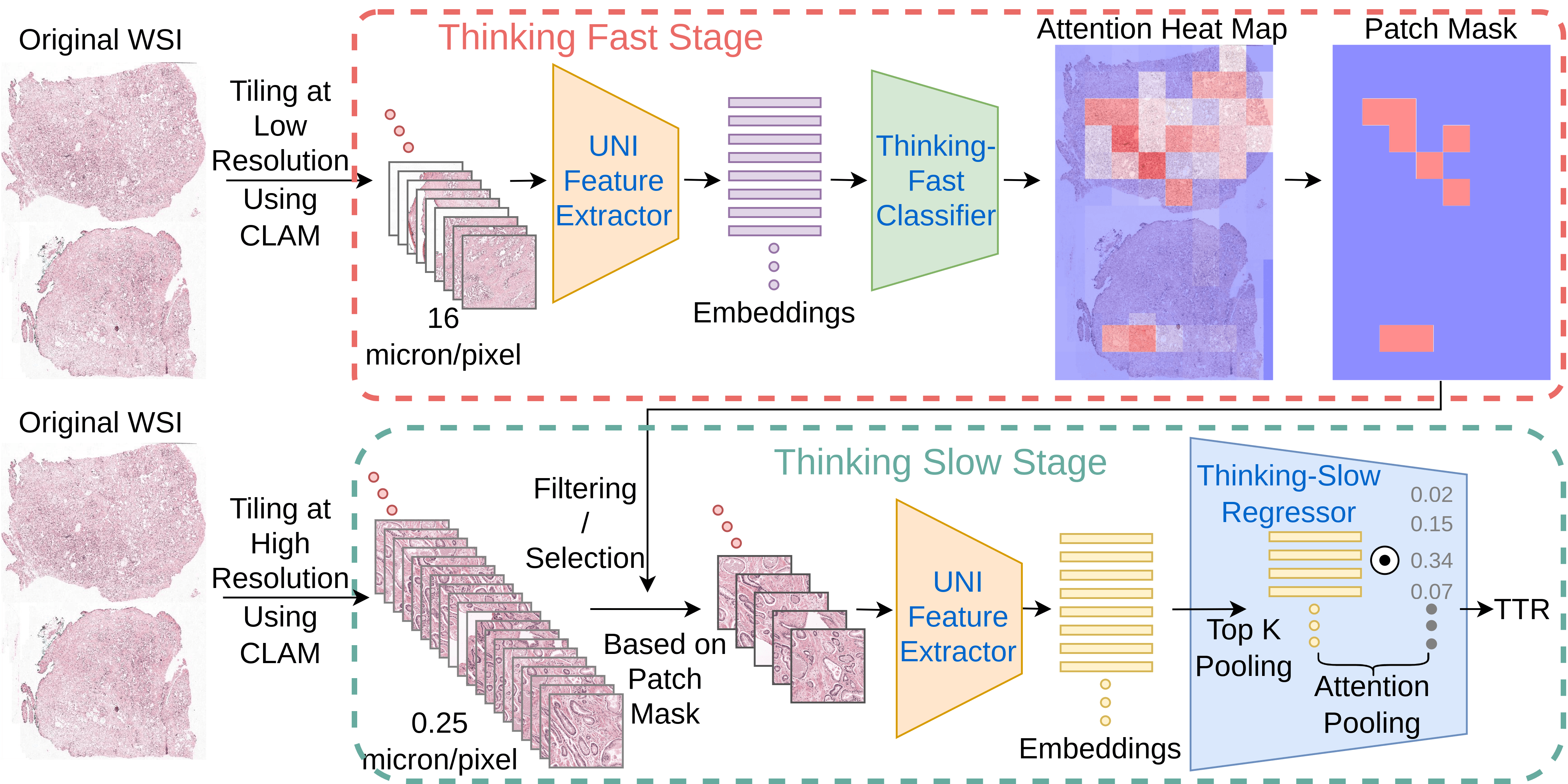} 
  \caption{Illustration of our two-stage ``Thinking fast \& slow'' approach. During ``thinking fast'', a patch mask is generated to rapidly localize relevant WSI patches. During ``thinking slow'' stage, top k and attention pooling are used for WSI TTR prediction.}
  \label{fig:pipeline}
\end{figure*}

Our proposed method consists of two MIL-based stages (Fig.~\ref{fig:pipeline}). The $1^{st}$ stage (``thinking fast'') targets classification at a low WSI resolution ($\approx16$mpp, $\approx0.625X$magnification), and the $2^{nd}$ stage (``thinking slow'') focuses on regression at a high resolution ($\approx0.25$mpp, $\approx40X$magnification). This approach targets improved patch sampling/pooling and inference efficiency. CLAM~\cite{lu2021data} was used for pre-processing (WSI patching and excluding background).% In the following sections, we first introduce the learning framework of MIL and then detail each stage's specifications, as well as the conjunction component.

    % \subsection{Data}
    %     Here you briefly describe the data used at the Leopard challenge.
    %\subsection{WSI with MIL}
    % Generally introduce the conception of MIL
    %In the setting of MIL, a WSI with a ground truth label $Y$ is termed a bag $I$ of many instances $\{p_1,p_2,...,p_N\}$, each of which is a smaller patch extracted from this WSI, without knowledge of their independent/localized ground truth $\{y_1,y_2,...,y_N\}$. These patches are embedded in a lower dimensional space $D$ through pre-trained feature extractors $F: f_n = F(p_n)$. The extracted embeddings $\{f_1,f_2,...,f_N\}$ are aggregated and/or pooled to represent the slide feature $e_n^D$ and used to make a final prediction $\hat{Y}$ for the WSI, through a fully connected layer $L: \hat{Y} = L(e_n^D)$.

    In the $1^{st}$ stage we extracted non-overlapping patches ($224\times224$), whereas patches in the $2^{nd}$ stage were of size $2048\times2048$ with 75\% overlap (step size=1024). These were embedded in lower dimensional spaces through UNI and used for classification (recurrence or not) in stage 1, and for TTR regression in stage 2.
    
    \subsection{Thinking Fast: MIL Classification at Low Resolution}\label{sec:stage_1}
    
    % This subsection starts with one paragraph to describe the coarser-level classification as the pre-selection process of patches for regression. The purpose of this process is to select presumably the most impactful and active areas of the WSI that might affect the prediction of the regression. 
    
    The $1^{st}$ stage intends to facilitate the rapid selection of WSI areas with the largest contribution to the TTR prediction, given a particular time threshold $T$. Its goal is reduced inference time and increased performance of the proposed approach. The recurrence of the WSI is defined as:
    \begin{equation}
        Y|_{t = T} = \begin{cases}
        0 &\text{if } \sum_{i = 1}^N{y_i} = 0 \\
        1 &\text{otherwise}
    \end{cases}
    \label{eq:classification}
    \end{equation}
    where $y_i$ is the prediction for the $i^{th}$ WSI patch and $Y|_{t = T}$ is the prediction for the WSI, at time threshold $T$. For this classification, we apply the CLAM-SB~\cite{lu2021data} model as the ``thinking fast'' classifier, which includes the patch loss and the cross-entropy for the WSI.
    % another paragraph describes the classification task that aims to predict the probability of the occurrence after surgery. The reasoning is that the higher the probability of cancer's occurrence manifests the higher risk of patient that develops cancer within a particular period of time (not sure if any reference I can find). Therefore, related image patches are more important to predict the occurrence time. 
    
    % This subsection ends with a description of loss function (with a math equation here) that was used for this classification (the clam-sb method), and a brief description of the classification task, including the output being a patch-wise probability map that helps determine/select/filter patches.
    After prediction, the probability of recurrence for each patch is generated and the top $m$ percent (up to 40\%) of patches with the highest attention scores will be assigned 1 in a mask, and 0 otherwise. This mask intends to filter out the less relevant tissue, in preparation for the second stage MIL process. 
    
    % Then in this paragraph, we describe how the patches (for the regression task) are selected by masking, and how this mask is generated from the probability map in the last paragraph. The mask is created by setting the top x percentage of patches in terms of prediction probability value as the foreground and the rest as the background.
    
    % The last paragraph talks about using UNI (may be a simple introduction) to extract the feature and as a bag (one WSI) for training.
    
    \subsection{Thinking Slow: MIL Regression at High Resolution}\label{sec:stage_2}
    Following the work of~\cite{katzman2018deepsurv}, the ``thinking slow'' stage is the regression task of predicting the patient risk $R$ for biochemical recurrence. This risk is inversely related to TTR. Thus, the output layer is described by a Cox Proportional Hazard~\cite{harrell2015cox} (CPH) layer, which is a single node and outputs the logarithmic risk $h(S)$ of a WSI feature embedding $S = \{f_1,f_2,...,f_N\}$. The WSI feature embeddings are extracted from patches selected by the mask of the ``thinking fast'' stage, after being pooled and aggregated for regression.
    
    In the CPH model, the risk $R(S) = e^{h(S)}$ is estimated by the linear function $\hat{h}_\beta(S) = \beta^T \cdot S$. In Cox regression, the weights $\beta$ are optimized by the Cox partial likelihood, which is defined as:
    \begin{equation}
    L(\beta) = \prod_{i:e_i = 1} \frac{e^{\hat{h}_\beta(S_i)}}{\sum_{j:R(t_i)} e^{\hat{h}_\beta(S_j)}} ,
    \label{eq:partial_likelihood}
    \end{equation}
    where $e_i$ is the event status (recurrence: 1, or not: 0) at follow up $t_i$ (in years), and $S_i$ is the WSI embedding. $R(t_i)$ indicates that the patient, whose input is the WSI, is still at risk of recurrence at time $t_i$. The optimization of Cox partial likelihood is equivalent to minimizing the following negative log partial likelihood function through re-parameterization:
    
    \begin{equation}
    l(\beta) = - \sum_{i:e_i = 1} ( \hat{h}_\beta(S_i) - \log \sum_{j:R(t_i)} e^{\hat{h}_\beta(S_j)} ),
    \label{eq:negative_log_partial_likelihood}
    \end{equation}    
    
    In our design, patch embeddings $\{f_1,f_2,...,f_N\}$ output their corresponding logarithmic risk $\{r_1,r_2,...,r_N\}$ through the Cox layer. Then, embeddings with top $k$ logarithmic risk are selected as $\{f_{top_1},f_{top_1},...,f_{top_k}\}$ (Fig.~\ref{fig:pipeline}). Among these embeddings, we define the pooling as a self-attention process~\cite{ilse2018attention}
    % \begin{align}
    % \begin{split}
    %     & S = \sum_{i = top_1} ^{top_k} a_i r_i, \\
    %     & a_i=\frac{\exp \{\mathbf{w}^{\top} \tanh (\mathbf{V}r_i^{\top})\}}{\sum_{j=top_1}^{top_k} \exp \{\mathbf{w}^{\top} \tanh \mathbf{V} r_j^{\top} \}},
    % \end{split}
    % \label{eq:topk_att}        
    % \end{align}

    \begin{equation}
        S \approx S_{top_k} = \sum_{i = top_1} ^{top_k} a_i r_i, \quad
        a_i=\frac{\exp \{\mathbf{w}^{\top} \tanh (\mathbf{V}r_i^{\top})\}}{\sum_{j=top_1}^{top_k} \exp \{\mathbf{w}^{\top} \tanh \mathbf{V} r_j^{\top} \}},
        \label{eq:topk_att}      
    \end{equation}
    where $ \mathbf{w}$ and $ \mathbf{V}$ are learnable parameters. $S_{top_k}$ is the $top_k$ embeddings weighted by attention pooling, designed to approximate the WSIs feature embeddings $S$ (Eq.~\ref{eq:partial_likelihood} \&~\ref{eq:negative_log_partial_likelihood}). $\tanh(\cdot)$ is an element-wise hyperbolic tangent function, introducing non-linearity. We approximate the TTR using $\exp(-1 \times \log R(S))$, since the logarithmic output risk $\log R(S)$ is inversely related to TTR.

    % requires the minimization of a log-loss function (add a math function here if any reference is required as the theoretic support, put it here). 
    
    % Second, we introduce the pooling method, where the top k predicted log risk is selected and then combined these selected features with a self-attention mechanism as follows:
    
    % ( which is topk plus attention), to achieve a better prediction. (topk and attention require references). We use exp(-1* log risk) as the predicted time. 

% \section{Experimental Design}
    \subsection{Model Training, Evaluation \& Selection}
    We used the Adam optimizer with a learning rate of $1\times 10^{-4}$. The weight decay was $1\times 10^{-5}$ and the dropout rate was 0.25. Models were trained and evaluated on NVIDIA A100 GPUs during model selection. Our source code is based on the CLAM platform and the tiffslide library.
    
    To select the best trained ``thinking fast'' model, we set up a 5-fold cross validation with a fixed test set and select the best fold as the model. The metric is the AUC of prediction on biochemical recurrence. For the ``thinking slow'' model, we use another 5x5-fold nested cross-validation without a fixed test set. In the outer fold, the hold-out set is used for validation of each inner fold. In each inner fold, the hold-out set is used to select the model for validation on the outer fold hold-out set, where the best inner hold-out validation loss is the criterion during training. The metric we used for model selection is the censored concordance index~\cite{uno2011c} ($Ci$) of the outer hold-out set. In our setting, 25 $Ci$ are calculated for one parameter setting (e.g., $top_k = 10$ and $m = 20\%$). In the experiments, we evaluated the model with combinations of $top_k = \{5, 10, 15, 20, 30, 40, 50\}$ and $m = \{5\%,10\%,15\%,20\%,25\%,30\%,35\%,40\%\}$. We select the model parameters by comparing the best mean and standard deviation ($\sigma$) of $Ci$.
    
    For the model submission to the LEOPARD challenge, we randomly split the data into a 10-fold cross-validation without a testing set and used the best model weights from each fold. The final prediction of TTR is calculated by averaging the predicted logarithmic risk from each set of model weights. We select model weights in each fold, based on the best hold-out validation loss, after 40 epochs. This 40-epoch threshold is set by calculating the zero-crossing epoch of the second derivative of the training loss curve to avoid under-training.
       
    % \subsection{Training Details}
\section{Results}

%\textcolor{blue}{Describe your experiments results on your own data split,  validation leaderboard, etc...}

    % \subsection{Nested Cross-validation Result \& LEOPARD Validation Result}
    % \begin{table}[h]
    %   \centering
    %   \caption{Internal Data Nested Cross-validation Result}
    %   \begin{tabular}{|c|c|c|c|}
    %     \hline
    %     Method & MAD-MIL Regression & AC-MIL Regression & \textbf{Ours} \\ \hline
    %     C-index Mean $\pm$ Std  &     &     & \textbf{0.733 $\pm$ 0.059}  \\ \hline
    %   \end{tabular}      
    %   \label{tab:ncv}
    % \end{table}
    
    For the internal data splits (Sec.~\ref{sec:material}), we selected the best performance with parameter settings $top_k = 10$ and $m = 20\%$. Our proposed approach yielded a mean C-index of 0.733 ($\sigma=0.059$) on our test data (i.e., the outer hold-out set), indicating superior performance compared to MAD-MIL~\cite{keshvarikhojasteh2024multi} (0.704 $\pm$ 0.058) and AC-MIL~\cite{zhang2023attention} (0.714 $\pm$ 0.056) with regression modifications.% (Table~\ref{tab:ncv}).
    % \subsection{LEOPARD Validation Result}     
    
    Our inference pipeline container submitted in the LEOPARD validation phase, yielded a C-index ($Ci$) of 0.603 ($Ci_{Radboud}=0.616, Ci_{external}=0.589$).    
    % \subsection{Ablations for Model Selection}
    
\begin{figure}[!ht]
      \centering
      \includegraphics[width=0.95\textwidth]{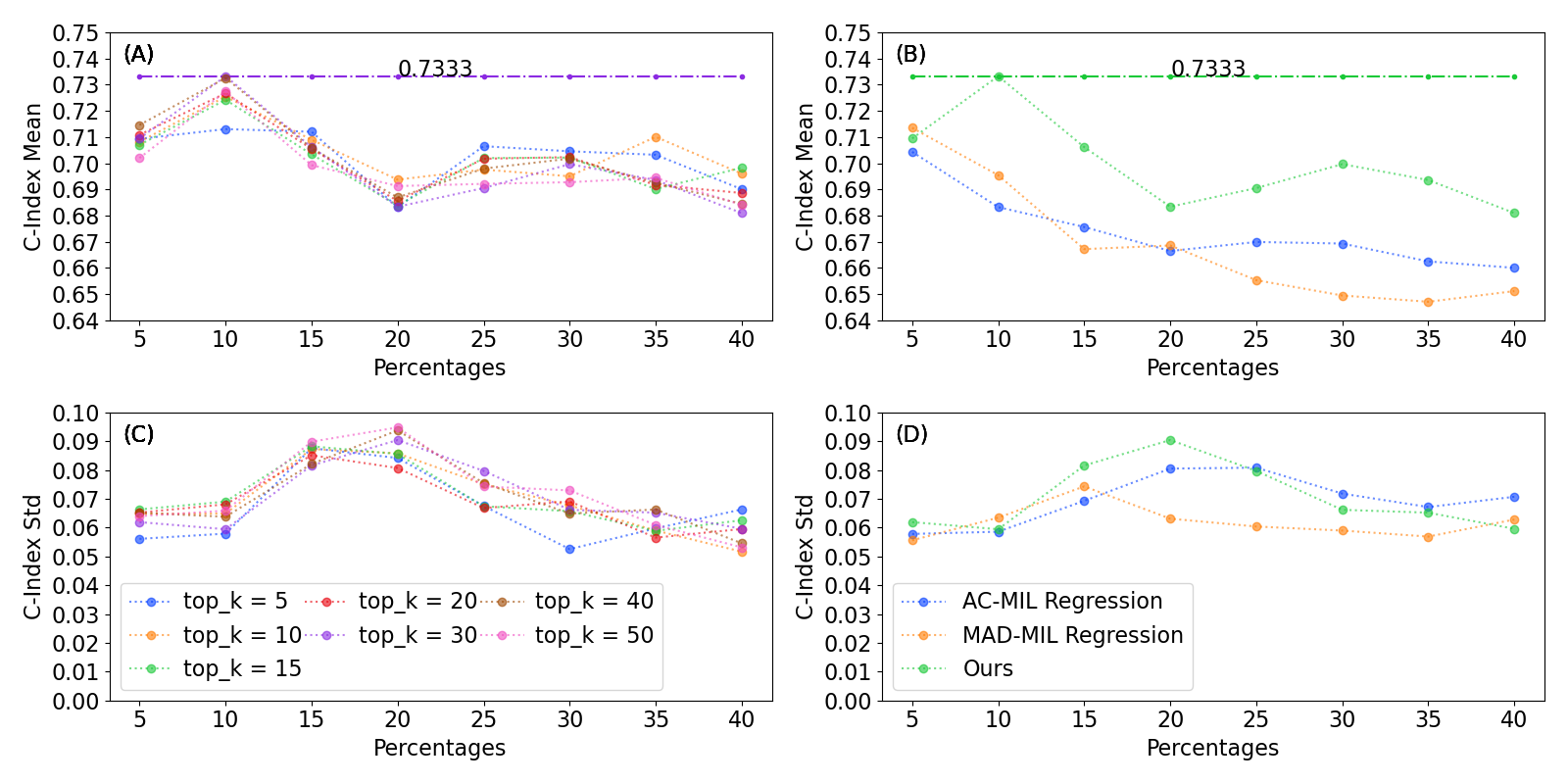}
      \caption{Ablation results for model selection. (A) and (C) show the comparison among different top k settings at different percentage of WSI ($m$). (B) and (D) compare the best parameter setting across MAD-MIL, AC-MIL, and our MIL method. y-axis in (A) and (B) represents mean $Ci$ values evaluated on the outer hold-out set and y-axis in (C) and (D) represents $\sigma_{Ci}$. The x-axis represents the percentage of patches used for the ``thinking slow'' stage.} 
      \label{fig: ncv}
    \end{figure}

    As shown in Fig.~\ref{fig: ncv} (A), we compare the results of different combinations of $top_k$ over $m$ percentage values (x axis). The upper plots show mean $Ci$ (y axis) of the outer hold-out set, while the lower plots show their corresponding standard deviation. The best overall result was observed for $top_k = 30$ and $m = 10\%$. We also observed that using a larger area of the WSI for regression does not always achieve better prediction, which in turn proves that a more relevant area of the WSI provides more accurate features for TTR regression and increases inference efficiency. This phenomenon can also be observed for the other two ablation methods, MAD-MIL and AC-MIL (Fig.~\ref{fig: ncv}(B)), where the selected method demonstrates a better regression prediction across almost all $m$ parameters when fixing other parameters. 

    % \subsection{Ablations in Model selection 2, methods}

\section{Interpretability} 
\begin{figure*}[h]
  \centering
  \includegraphics[width=0.95\textwidth]{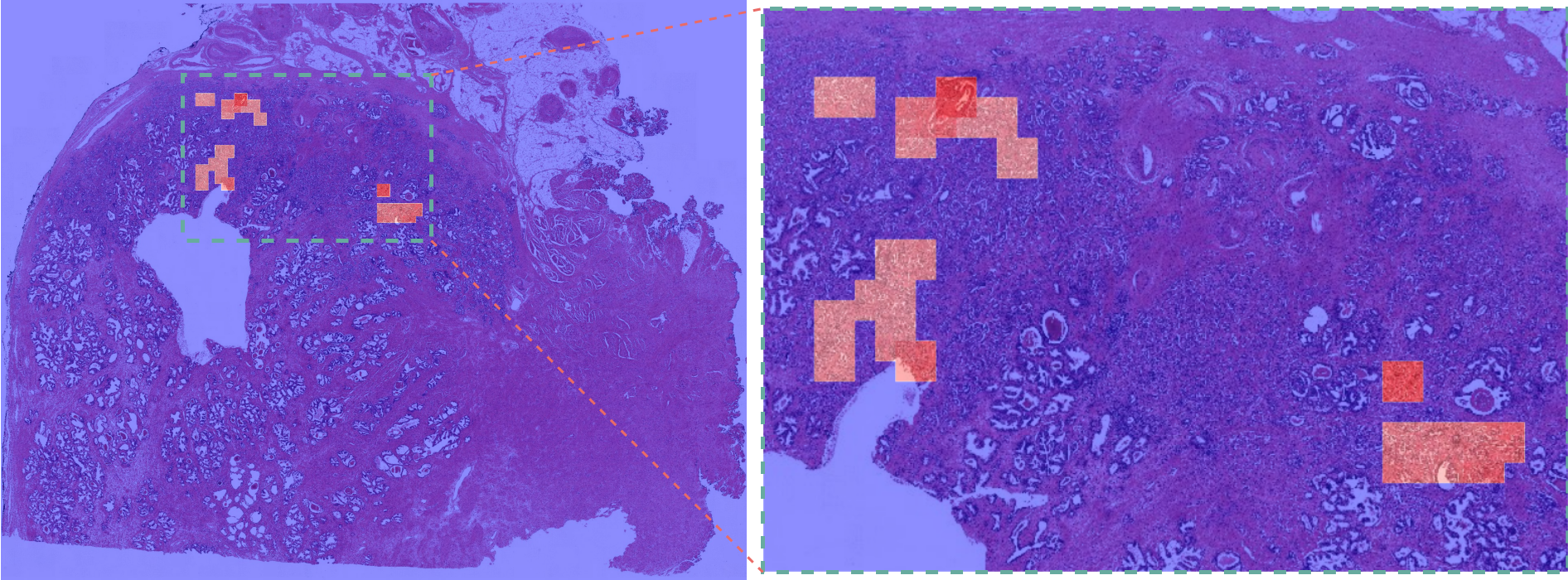} 
  \caption{One example shows interpretability for our model. Both images are overlays of the attention heat map on the tissue. The image on the right is a zoom-in for the selected area on the left. Higher attention scores are in more red-covered area.}
  \label{fig:interpretability}
\end{figure*}

% \textcolor{blue}{Describe if your method provides any interpretability of the scores.}
In the first stage, the patch selection criterion is the highest attention score (Fig.~\ref{fig:pipeline}) of the attention map, which serves as an interpretability visualization for previous classification works. It show the most attentive area for the stage one classfication. Shown in Fig.~\ref{fig:interpretability}, in our second stage, the attention scores are sparsely distributed on the WSI since only a small portion of patches (10\%) are selected. Those color-highlited area also shows the most attentive region for TTR regression. In general, our method leverages the attention mechanism, but further clinical interpretablity requires to be evaluated from clinicians/pathologists.% In the selected patches, high attention scores (in orange and red) are placed on top of highly unhealthy tissues, as suggested by the pathologist. 

\section{Discussion}
% \textcolor{blue}{Summarize and discuss your findings, limitations, and areas of future work.}

In this study, we proposed to leverage MIL through a two-stage ``thinking fast \& slow'' strategy for the TTR regression. The first ``thinking fast'' stage aims to find the most relevant area of the WSI to the biochemical recurrence and the second ``thinking slow'' stage leverages higher resolution patches to predict the TTR. In the ablation result, we have shown that an improved prediction can be achieved by focusing on a more relevant area of the WSI along with an improved prediction efficiency. We also showed that the regression is affected by areas of attention which contain cancerous tissues. The limitation of our method is from the CPH model, which focuses on the risk prediction, not the real TTR. In the future, we will extend our work to other tumor types.
\section{Code Link}
 The source code of our inference pipeline is available at \url{https://github.com/yousuhang/IU-ComPath-LeoPard}.
%
% ---- Bibliography ----
%
% BibTeX users should specify bibliography style 'splncs04'.
% References will then be sorted and formatted in the correct style.
%
% \bibliographystyle{splncs04}
% \bibliography{mybibliography}
%
\newpage
\bibliographystyle{splncs04}
\bibliography{ref}
\end{document}